# Safety-quantifiable Line Feature-based Monocular Visual Localization with 3D Prior Map

Xi Zheng, Weisong Wen*, Li-Ta Hsu*

*Abstract*—Accurate and safety-quantifiable localization is of great significance for safety-critical autonomous systems, such as unmanned ground vehicles (UGV) and unmanned aerial vehicles (UAV). The visual odometry-based method can provide accurate positioning in a short period but is subjected to drift over time. Moreover, the quantification of the safety of the localization solution (the error is bounded by a certain value) is still a challenge. To fill the gaps, this paper proposes a safety-quantifiable line feature-based visual localization method with a prior map. The visual-inertial odometry provides a high-frequency local pose estimation which serves as the initial guess for the visual localization. By obtaining a visual line feature pair association, a foot point-based constraint is proposed to construct the cost function between the 2D lines extracted from the real-time image and the 3D lines extracted from the high-precision prior 3D point cloud map. Moreover, a global navigation satellite systems (GNSS) receiver autonomous integrity monitoring (RAIM) inspired method is employed to quantify the safety of the derived localization solution. Among that, an outlier rejection (also well-known as fault detection and exclusion) strategy is employed via the weighted sum of squares residual with a Chi-squared probability distribution. A protection level (PL) scheme considering multiple outliers is derived and utilized to quantify the potential error bound of the localization solution in both position and rotation domains. The effectiveness of the proposed safety-quantifiable localization system is verified using the datasets collected in the UAV indoor and UGV outdoor environments.

*Index Terms*—Safety quantification, State estimation, Visual localization, Prior map, Outlier rejection, Protection level.

## I. INTRODUCTION

Accurate, cost-effective, and reliable localization is of great importance for the realization of safety-critical autonomous systems, such as unmanned ground vehicles (UGV) [1, 2] and unmanned aerial vehicles (UAV) [3]. The global navigation satellite system (GNSS) [4] is widely used for providing globally-referenced positioning for autonomous systems with navigation requirements. However, positioning accuracy is severely degraded in highly-urbanized cities such as Hong Kong, due to signal reflection caused by static buildings leading to the notorious non-light-of-sight (NLOS) receptions [5] and multipath effects, so-called GNSS outlier measurements.

Instead of relying on the GNSS, the prior map-based light detection and ranging (LiDAR) localization methods [6, 7] attracted lots of attention due to their robustness and accuracy, where the matching between the real-time point clouds and the prior map was employed to estimate the location of the system within the map. However, the consumption of computational resources in point cloud registration [8] and the high-cost LiDAR sensor limit the massive deployment of these localization solutions [9]. Recently, the 2D visual-based localization in a 3D prior map attracted lots of attention [10-12]. But the visual localization within the prior map is still challenging, due to the pattern difference between the 2D pixel texture measurements from the visual image and 3D point cloud information from the prior map [13]. Moreover, reliably quantifying the safety of the derived localization solution is significant for safety-critical autonomous systems. The navigation solution is certified as safe only when its potential error is quantifiable. However, the safety quantification of the map-based localization method is still an open question.

### A. Visual Localization with Prior Point Cloud Map

To fill the gap between the 2D image and 3D point cloud, some researchers transform data into the same dimensional spaces for the association. For example, the photometry matching-based methods proposed to project 3D points into synthetic images and match them with a camera intensity image by maximizing the normalized mutual information [14] or minimizing the normalized information distance [15]. However, the nonlinear projection transformation from points to images would be challenged by the discreteness of the point cloud and limit the localization accuracy. Besides, the geometry matching-based methods employed bundle adjustment (BA) [16] to reconstruct a point cloud, and then align the two point clouds with registration methods, such as the iterative closest point (ICP) [8, 17], and normal distribution transform (NDT) [18, 19]. The hybrid of photometric and geometric methods calculated two types of depth images from point clouds and a stereo camera, respectively, and then compared them for camera localization [11, 20]. However, large-scale BA operation is time-consuming, and the point cloud reconstructed based on image features is usually sparse. Meanwhile, the range of the depth recovery from the stereo camera is limited because of the requirement of dense texture information.

The other research stream [10, 21] proposed to directly match the 2D representative features from the image with the 3D point cloud from the prior map by estimating the scale and the pose of the system simultaneously. However, these methods relied heavily on the initial guess of the pose of the system during the 2D-to-3D association. Moreover, those methods only exploited the potential of corner features, and the geometry information (such as lines) from the environments is not fully explored. On the other hand, some learning-based networks conduct image-to-point cloud registration by training various datasets, but these methods ultimately suffer from the dependence on labeled datasets and poor generality [22, 23].

Interestingly, the line feature was utilized by visual-simultaneously and mapping (VSLAM) communities [24, 25] where the lines are employed as the additional feature correspondence to estimate the motion of the system. In some low-textured areas (e.g. the conventional corner features are limited), line segments containing geometrical structure information are particularly effective [26]. Furthermore, lines usually have more stable constraints than points in both 3D space and 2D images, as a line segment consists of multiple points [27, 28]. The results [24, 25] revealed that the line features could effectively exploit the potential of the geometry information within the environment. Considering point clouds with salient structural information and inspired by the potential of the prior map, the work in [29] proposed a line feature-based visual localization method where the 2D line feature was associated with the 3D lines from the prior point cloud map aided by the high-frequency initial guess of the pose from the visual-inertial system (VINS) [30]. However, the work in [29] constructed the constraints of the line correspondence based on the line-to-line distance which fails to fully explore the geometry correlation of the line features. *As an extension, this paper proposes an improved line feature-based point-to-point constraint model aided by a carefully designed line correspondence selection strategy to achieve reliable visual localization within the prior map.*

### B. Safety Quantification of Localization Solution

Reliably quantifying the safety of the derived localization solution is significant for safety-critical autonomous systems before its massive production and deployment. The navigation solution is certified as safe only when its potential error is quantifiable [31] and could be monitored simultaneously. Unfortunately, the safety quantification of the map-based localization solution is not effectively studied, which still has a big gap in the front. Specifically, the potential error of the localization solution is mainly caused by two parts: (1) *The random Gaussian noise*: the Gaussian noise associated with the raw observation measurements could lead to potential localization uncertainty [32]. (2) *The unmodelled bias from outliers*: the unmodelled outlier measurements with bias can lead to additional uncertainty in the localization solution [33].

Scientifically speaking, the key to quantifying the safety of the localization solution is to reliably account for the uncertainty arising from those two components. In visual localization, several outlier rejection algorithms have existed. The popular methods are random sample consensus (RANSAC) and its variants [34, 35]. However, the RANSAC algorithm relies on the selection of the kernel parameter which can be slow and fail in the presence of high outlier rates [36]. On the other hand, the global optimization methods based on branch-and-bound (BnB) and mixed integer programming strategies try to make the global outlier searching problem tractable and faster [13, 37]. However, these methods could lead to a high computational load and a satisfactory initial guess of the state is required. Another method for outlier detection is the parity space approach (PSA), which projects the measurements into a designed parity space, where the outliers can be highlighted and detected [38]. However, due to the misdetection and omission of outliers, and the random Gaussian noise contained in the observations, the safety of the visual navigation system still cannot be guaranteed.

Interestingly, the state-of-the-art GNSS receiver autonomous integrity monitoring (RAIM) theory quantifies the safety of the satellite positioning solution by systematically considering the uncertainty from the satellite measurements and the unmodelled bias from outliers [32, 33]. In GNSS, the RAIM theory starts with the fault detection and exclusion (FDE) process which aims to remove the potential outlier satellite measurements. Then the safety of the positioning solution is quantified based on the survived satellite measurements by calculating a statistical value, so-called the protection level (PL), which aims to account for the maximal potential error of the positioning solution caused by the Gaussian noise and the outliers that survived the FDE [32, 33]. Initially, the calculation of PL was mainly based on the assumption that only one outlier is left in the observations [32], and then the work proposed in this paper [33] extended the basic assumption from the single outlier to multiple outliers.

Inspired by the RAIM theory in the GNSS field, Li et al. drew on RAIM to monitor the integrity of visual localization based on ORB-SLAM2 [39] and determined an approximate upper bound of the state estimation error by calculating the protection level [39, 40]. Dr. Zhu proposed a preliminary visual positioning pipeline with integrity monitoring and focused on the error model of visual pixel features for fault detection and elimination, but this pipeline has not been systematically verified [41, 42]. Meanwhile, these methods mainly rely on the single fault assumption both in FDE and safety quantification which limits its generality for the cases with multiple faults. Moreover, the RAIM theory in GNSS only considered safety quantification in the positioning domain but the safety quantification in the orientation is also equally important for autonomous systems (e.g, UAV), which is not fully investigated. *In this paper, a RAIM-inspired safety quantification method is proposed to enable the certification of safety under the multiple-outlier assumption, and in both positioning and orientation domains.*

### C. Key Contributions of This Paper

This paper proposes a safety-quantifiable line feature-based visual localization method with a prior map. The main contributions of this paper are summarized as follows:

1) *Pixel-level point-to-point constraint model for visual localization*: By transforming the 2D-3D line association-based optimization problem from minimizing the line-to-line distance model into a point-to-point reprojection model, a new reliable pixel-level constraint model based on line features is proposed for reliable localization within the prior map.

2) *Safety quantification considering multiple faults*: A RAIM-inspired safety quantification pipeline is proposed which accounts for the localization safety both for the positioning and orientation domains. This paper systematically derives the safety quantification process by FDE and PL estimation.

3) *A comprehensive framework for safety quantifiable visual localization:* This is the first proposed integral framework including visual localization and safety quantification for autonomous systems. The effectiveness of the proposed pipeline is verified with challenging datasets collected by both UAV and UGV systems. Moreover, a detailed analysis of the localization safety towards the line features distribution is presented.

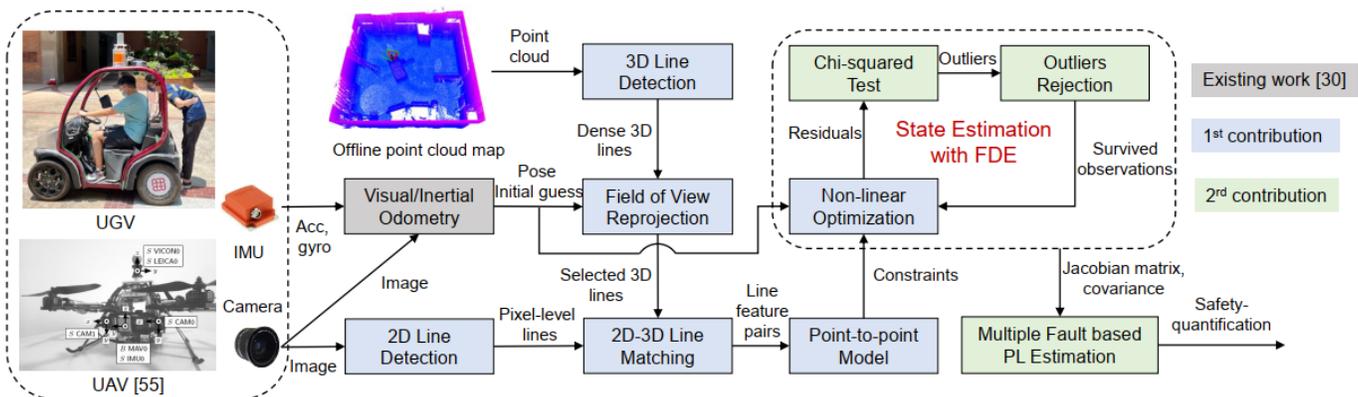

Fig. 1: The framework of the proposed system. The 2D, 3D lines are detected in camera frame online and prior 3D point cloud map offline, respectively (see Section III-A). After giving the state initial guess by VINS [30], the two types of lines are matched as correspondences (see Section III-A), which are observations for the optimization model (see Section III-B). Based on the residuals of optimization, a Chi-squared test is utilized to reject outliers iteratively (see Section IV-B, fault detection and exclude (FDE), i.e., outlier rejection). When the test is passed, a protection level (see Section IV-C) is calculated to quantify the safety of state estimation.

The rest of this paper is organized as follows. The overview of the proposed system pipeline is given in Section II. The methodology of the proposed visual localization with a prior map based on line feature correspondences is presented in Section III. In Section IV, the derivation of the safety quantification is presented including the FDE, and the protection level estimation considering multiple faults is presented. Experimental results and discussions are provided in Section V. Section VI concludes this paper and outlines the future research directions. The *Appendix* supplements the details of the Jacobian matrix used in Section III.

## II. OVERVIEW OF THE PROPOSED METHOD

The proposed system pipeline is shown in Fig. 1. The system mainly includes two parts, the visual localization module, and the safety quantification module. Firstly, a high-precision 3D point cloud as a prior map is generated in advance, in which all 3D lines are extracted and stored offline [43]. Then the 3D lines are projected into an image frame based on the initial field of view (FoV) of the camera that is provided based on the pose estimation from the VINS [30]. On the other hand, the images captured by an onboard camera are used to detect 2D lines in real-time (Section III-A). The two types of lines are matched into line feature correspondences, which are utilized as the predictions and measurements of the state optimization model for absolute visual localization (Section III-B) in the prior map. Secondly, a safety quantification is implemented based on reliable FDE and protection level calculation. Specifically, a Chi-squared test is applied to detect outliers based on the weighted residual iteratively (Section IV-B). After removing the detected outliers, the survived observations are re-input into the optimization model for state estimation until the Chi-squared test passes. Then the protection level is estimated based on the multiple fault assumption by quantitatively evaluating the safety of the visual localization solution (Section IV-C).

The notations are defined as follows. $(\cdot)^w$ is the world frame, and the origin coincides with the prior map starting point. $(\cdot)^b$ is the body frame, and the origin is defined on the inertial measurement unit (IMU). $(\cdot)^c$ is the camera frame. The translation vector and rotation matrix are represented by $R$ and $t$ respectively, and also by their corresponding Lie algebra $\xi$ [44]. $R_w^b, t_w^b$ means the transformation from the world frame to the body frame, $R_b^c, t_b^c$ is from the body frame to the camera frame. $b_k$ is the subscript of the body frame in $k^{th}$ image. The state variables in this paper are the pose of UGV or UAV at world frame while taking the $k^{th}$ image, expressed by $R_{b_k}^w$ and $t_{b_k}^w$. $P(X, Y, Z)$ is a 3D point and $p(\mu, v)$ means a 2D image pixel point. The notation of Chi-Squared distribution is $\chi^2$.

## III. STATE ESTIMATION: VISUAL LOCALIZATION WITH PRIOR MAP

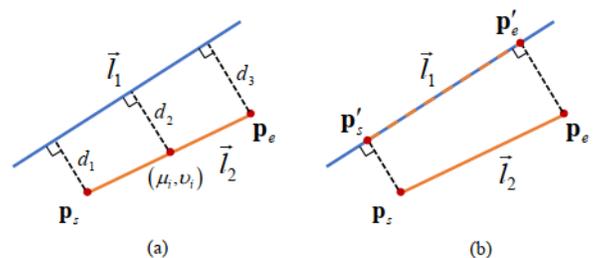

Fig. 2: An illustration of line similarity criteria. (a) the line distance is the sum lengths of the dashed lines, denoted by $\sum d_i$. (b) the line overlap rate refers to the percentage of the orange dashed line on the blue line.

### A. 2D-3D Lines Matching

In this paper, the 2D lines are detected by the LSD method [45] and are indicated by two endpoints $(\mathbf{p}_s, \mathbf{p}_e)$ in 2D image. The line features existing in 3D point clouds are extracted by a segment-based 3D line detection method. In the method, the 3D points are segmented by planes and projected into a virtual image, then the 2D lines detected on this image are re-projected into 3D space to get line segments [43]. Once the 3D lines are obtained, the dense point cloud can be replaced by line segments that are represented by two 3D endpoints ($\mathbf{P}_s, \mathbf{P}_e$), and only these lines need to be loaded during visual localization.

To get line correspondences, we first project the 3D lines into the camera imaging plane according to the initial pose of the camera from VINS. Then, these lines are matched with the 2D lines captured from the actual world in this image depending on the defined similarity criteria. The criteria in this paper are mainly determined by the distance, angle, and overlap between two line segments [13, 29]. Specifically, the general function of a 2D line $l$ is

$$Ax + By + C = 0, \quad (1)$$

the distance $D$ is defined as the sum of distances from $n$ equally interval sampling points $(\mu_i, v_i), i = 1 \cdots n$ (containing endpoints) on one line segment to the other (shown in Fig. 2 (a)), the equation is,

$$D = \sum_{i=1}^{n} d_i = \sum_{i=1}^{n} \frac{|A\mu_i + Bv_i + C|}{\sqrt{A^2 + B^2}}. \quad (2)$$

The angle $\theta$ of two lines $l_1, l_2$ is

$$\theta = \angle(\vec{l_1}, \vec{l_2}), \quad (3)$$

and the overlap $v$ is shown in Fig. 2 (b),

$$v = \frac{|\overrightarrow{\mathbf{P}'_s \mathbf{P}'_e}|}{|\vec{l_1}|}. \quad (4)$$

By assigning appropriate thresholds which are experimentally determined, the algorithm can obtain line correspondences that meet requirements. Fig. 3 (a) shows the line detection result at one camera frame, and (b) is the line matching result on this frame.

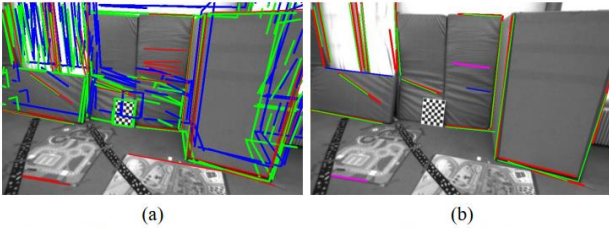

Fig. 3: (a) The red lines in the image are the detected 2D lines by FLD method; the blue lines are the projected lines from 3D map by VIO FoV, and the FoV of green lines is based on our state estimation result (More details can be seen in Section III-B). (b) The red and green line pairs are the valid matching association, the blue and purple lines are the rejected outliers (see Section IV-B).

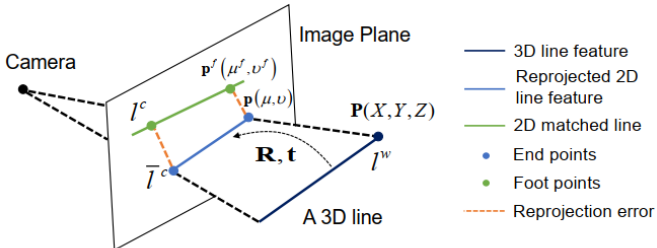

Fig. 4: An illustration of line reprojection error from the 3D prior map to the 2D image. A 3D point $\mathbf{P}(X,Y,Z)$ on $l^w$ is projected into a camera image with $\mathbf{p}(\mu, \nu)$ based on the transformation $\mathbf{R}, \mathbf{t}$. The foot point $\mathbf{p}^f(\mu^f, \nu^f)$ of $\mathbf{p}(\mu, \nu)$ on $l^c$ is found to build a pixel cost function that is be minimized.

### B. Point-to-point Optimization Model

Generally, the cost function among line features can be built based on the sum distance between line correspondences. In this paper, the optimization model is transformed from minimizing the distance of lines to point reprojection error [44].

Assuming there is a line pair $l_1, l_2$, the general function of $l_1$ is shown in (1), and the distance between them can be calculated by the sum distance of the points on $l_2$ from $l_1$, which is consistent with equation (2). Take one point $p(\mu, v)$ on $l_2$ as an example, the distance from $p$ to $l_1$ is equivalent to the distance from the point $p$ to its vertical foot point $p^f(\mu^f, v^f)$ on $l_1$. This point-to-point distance is essentially the same as the reprojection error in visual pose estimation, both of them are used to represent the difference between the coordinates of two-pixel points. Based on this, the objective function can be built by minimizing the square sum of the point pair error,

$$\mathbf{F} = \min \sum \left\| \mathbf{p}^f - \mathbf{p} \right\|_{\mathbf{Q}}^2, \quad (5)$$

where $p^f$ and $p$ are the measurement and prediction of this function, respectively. The $\mathbf{Q}$ is the covariance of the observations and is experimentally determined. The coordinate of the foot point $p^f(\mu^f, v^f)$ can be computed as

$$\mu^f = \frac{B^2\mu - AB\nu - AC}{A^2 + B^2}, \nu^f = \frac{A^2\nu - AB\mu - BC}{A^2 + B^2}. \quad (6)$$

The specific process of constructing an optimization model is illustrated in Fig. 4. Suppose there is a 3D line segment $l^w$ in space at the time while the $k^{th}$ image taken by the camera, and one of its endpoints is $P(X, Y, Z)$. The transformation of the camera during this time from the world frame is $T_w^c = [R|t]$ (corresponding to Lie algebra $\xi$ [46]), the 3D line $l^w$ is projected into the image plane of the camera shown as $\bar{l}^c$, which is matched with the line segment $l^c$ in this image. Based on the pre-calibrated extrinsic matrix $T_w^c$, this projection model can be expressed as

$$s \begin{bmatrix} \mu \\ \nu \\ 1 \end{bmatrix} = \mathbf{KT} \begin{bmatrix} X \\ Y \\ Z \\ 1 \end{bmatrix} = \mathbf{K} \exp(\xi^{\wedge}) \mathbf{P}, \quad (7)$$

where $p(\mu, v)$ is the pixel location of the endpoint $P(X, Y, Z)$ in an image plane. s is the scale and $\mathbf{K}$ is the intrinsic matrix of the camera [47]. The symbol $\wedge$ is a vector to skew-symmetric conversion. Combined with the state variables we defined in Section II, the transformation $T_w^c$ can be decomposed as

$$\begin{aligned} \mathbf{R}_w^c &= \mathbf{R}_{b_k}^c \mathbf{R}_w^{b_k} \\ \mathbf{t}_w^c &= \mathbf{R}_{b_k}^c \mathbf{t}_w^{b_k} + \mathbf{t}_{b_k}^c \end{aligned}. \quad (8)$$

where $R_{b_k}^c$ and $t_{b_k}^c$ refer to the transformation between the camera and IMU at the $k^{th}$ image, which are invariant parameters because the camera and IMU are fixedly mounted on autonomous robots, so we have $T_{b_k}^c = T_{b_0}^c = T_b^c$, and this transformation can be obtained by the joint camera-IMU calibration [48]. In our system, the initial guess of states $T_w^{b_k}$ is provided by VINS [30].

In the image plane, after obtaining the prediction $p(\mu, v)$ based on (7), the measurement $p^f(\mu^f, v^f)$ on the matched line pair $l^c$ can be computed by (6). So the error function depended on the point-to-point reprojection error can be shown as

$$\mathbf{e} = \mathbf{p}^f - \mathbf{p} = (\mu^f - \mu)^2 + (\nu^f - \nu)^2. \tag{9}$$

Therefore, the cost function (5) can be rewritten by

$$\xi^* = \min_{\xi} \sum_{k=1}^{N} \left\| \mathbf{p}_k^f(\mu^f(\xi), \nu^f(\xi)) - \frac{1}{s_k} (\mathbf{K} \exp{(\xi^\wedge)} \mathbf{P}_k) \right\|_{\mathbf{Q}}^2, \tag{10}$$

where N is the number of line pairs on the current image. The proposed method selects two endpoints from each set of pairs as observations for the optimization model. Moreover, the measurements $p^f$ is a variable related to the state $\xi$.

### C. Optimization Solver

For the non-linear optimization problem in this paper, the Gauss-Newton (GN) method is employed and is implemented by the Ceres solver [49]. First, the error function is linearized by Taylor expansion about x shown as

$$\mathbf{e}(\mathbf{x} + \Delta \mathbf{x}) \approx \mathbf{e}(\mathbf{x}) + \mathbf{J}^\top \Delta \mathbf{x}. \tag{11}$$

where x is the state variable, and J is the Jacobian matrix. If the Jacobian matrix is known, by giving the initial guess of states x, the optimization increment $\Delta x$ can be obtained by

$$\mathbf{H} \Delta \mathbf{x} = \mathbf{g} \tag{12}$$

where H is $J^T J$, g is $-J^T e$ [50]. Then the states are updated by $x_{i+1} = x_i + \Delta x$, until the algorithm converges, where $i$ is the iterative index.

In terms of the Jacobian matrix in this paper, the proposed method utilizes the perturbation model in lie algebra to derive it [44]. Based on the chain rule and (9), the derivation of each error term with respect to the state variable can be calculated by

$$\mathbf{J} = \frac{\partial \mathbf{e}}{\partial \delta \xi} = \lim_{\delta \xi \to 0} \frac{\mathbf{e}(\delta \xi \oplus \xi) - \mathbf{e}(\xi)}{\delta \xi} = \frac{\partial \mathbf{e}}{\partial \mathbf{p}} \frac{\partial \mathbf{p}}{\partial \mathbf{P}'} \frac{\partial \mathbf{P}'}{\partial \delta \xi}. \tag{13}$$

Here $\oplus$ refers to the disturbance left multiplication in Lie algebra, $\delta \xi$ is the perturbation value. The $P'$ is a point in the camera frame, which is transformed by the 3D space point $P$ in the world frame, referring to equation (7), we have

$$\mathbf{P}' = (\mathbf{TP})_{1:3} = \exp{(\xi^\wedge)} \mathbf{P}_{1:3} = [X', Y', Z']^\top. \tag{14}$$

The specific calculation procedure for the three components of the Jacobian matrix in (13) can be found in Appendix A. Therefore, the localization problem could be solved iteratively based on the devised Jacobian matrices.

## IV. SAFETY QUANTIFICATION OF VISUAL LOCALIZATION WITH PRIOR MAP

### A. Observation Model

Considering the state estimation problem in the previous section, a general observation equation could be constructed as follows,

$$\mathbf{z} = \mathbf{h}(\mathbf{x}) + \epsilon, \tag{15}$$

where **x** is the state variable and $\varepsilon$ is the noise item. z is the measurements, corresponding to the foot point $p^f$. $h(\cdot)$ is a nonlinear measurement model and corresponds to (7) in Section III, which can be linearized by a first-order Taylor expansion as shown in Section III-C to get an approximated linear model,

$$\hat{\mathbf{z}} = \mathbf{J} \Delta \mathbf{x} + \epsilon \tag{16}$$

where J is the Jacobian matrix. $\hat{z} = z - h(x_0)$ is the shifted measurement according to the operating point $x_0$ (initial guess of state variables). Here we assume that the error introduced by the linearization is negligible. As in the case of visual localization, the frame frequency is high and the movement between successive frames is small, making the operating point close to the state convergence value [51].

Generally, the noise item $\varepsilon$ is assumed to be a Gaussian distribution with zero mean and a known covariance matrix Q,

$$\epsilon \sim \mathcal{N}(0, \mathbf{Q}). \tag{17}$$

However, in this section, a fixed bias $b$ is added to the error model as

$$\epsilon \sim \mathcal{N}(\mathbf{b}, \mathbf{Q}). \tag{18}$$

The bias $b$ represents another source of error, here called outliers, the details will be discussed in the following section.

For the linear problem in (16), we can solve it by the weighted least squares method, and the solution is

$$\Delta \hat{\mathbf{x}} = (\mathbf{J}^\top \mathbf{W} \mathbf{J})^{-1} \mathbf{J}^\top \mathbf{W} \hat{\mathbf{z}}, \tag{19}$$

where W is the weighted matrix or information matrix and is the inverse of noise covariance matrix Q,

$$\mathbf{W} = \mathbf{Q}^{-1}. \tag{20}$$

Note that the Jacobian matrix J could be updated in each iteration.

### B. Outlier Rejection

Referring to Zhu's work [41], we could briefly classify the errors associated with the raw visual measurements (line features in this paper) in the proposed framework as follows. The first category is the photometric noise caused by camera calibration error, camera over-exposure, and motion blur. These errors will affect the detection accuracy of line features and are expressed as the covariance matrix Q of the error model in (18). The second category is the fixed estimation deviations introduced by outliers, and we use the bias $b$ in (18) to represent the effect of the outliers on the state estimation. However, it is undeniable that the presence of outliers is associated with the first category, as this category can cause feature misdetection and mismatching.

Due to the existence of noise and bias in measurements, the results of state estimation could contain residual $\varepsilon$. Combining (16) and (19), the residual $\varepsilon$ can be written as

$$\begin{aligned} \varepsilon &= \hat{\mathbf{z}} - \mathbf{J} \Delta \hat{\mathbf{x}} = (\mathbf{I} - \mathbf{P}) \hat{\mathbf{z}} \\ \mathbf{P} &= \mathbf{J}(\mathbf{J}^\top \mathbf{W} \mathbf{J})^{-1} \mathbf{J}^\top \mathbf{W} \end{aligned}. \tag{21}$$

In the actual calculation, the residual is obtained from the optimization solver in the previous section.

To evaluate the quality of measurements, a weighted sum of the squared errors (WSSE) [32] based on the residual is defined as

$$\varepsilon^\top \mathbf{W} \varepsilon = [(\mathbf{I} - \mathbf{P})] \hat{\mathbf{z}}^\top \mathbf{W} [(\mathbf{I} - \mathbf{P}) \hat{\mathbf{z}}], \tag{22}$$

which can be further simplified as

$$\begin{aligned} \varepsilon^\top \mathbf{W} \varepsilon &\Rightarrow \hat{\mathbf{z}}^\top \mathbf{W} (\mathbf{I} - \mathbf{P}) \hat{\mathbf{z}} = \hat{\mathbf{z}}^\top \mathbf{S} \hat{\mathbf{z}} \\ \mathbf{S} &= \mathbf{W} (\mathbf{I} - \mathbf{J}(\mathbf{J}^\top \mathbf{W} \mathbf{J})^{-1} \mathbf{J}^\top \mathbf{W}) \end{aligned}. \tag{23}$$

Next, a Chi-squared test for WSSE is used to determine whether there are outliers in the measurements:

$$\hat{\mathbf{z}}^\top \mathbf{S}\hat{\mathbf{z}} > \chi^2_{1-\alpha}(n-m) \equiv \mathbf{TD}, \quad (24)$$

where, $\chi^2_{1-\alpha}(n-m)$ is the $1-\alpha$ quantile of the central Chi-squared distribution with $(n-m)$ degrees of freedom (DoF) [33]. $n$ is the number of measurements, $m$ is the number of state variables, $\alpha$ is the is the *probability of false alarm* ($P_{fa}$) and is set to 0.05 in this paper. $\mathbf{TD}$ is a threshold that can be obtained by checking the $\chi^2$ distribution table according to the DoF and confidence $1-\alpha$. If

$$\hat{\mathbf{z}}^\top \mathbf{S}\hat{\mathbf{z}} > \mathbf{TD}, \quad (25)$$

the probability of outliers existing in the measurements is $1-\alpha$ (95%).

Because of the properties of the $\chi^2$ distribution and the error model (18), outliers in the measurements can be rejected by (24). If the bias $\mathbf{b}$ in (18) is not equal to 0, the WSSE will not fit a central $\chi^2$ distribution but a non-central distribution. The non-centrality parameter is $\lambda \equiv b^T S b$ [33], which is shown as

$$E(\hat{\mathbf{z}}^\top \mathbf{S}\hat{\mathbf{z}}) - (n-m) = b^\top \mathbf{S} b. \quad (26)$$

Therefore, the central Chi-squared distribution in (24) cannot be satisfied only if all outliers are theoretically removed from the observations (i.e. the bias in the error model is equal to 0). In this way, a greedy algorithm [40] can be applied to iteratively exclude the observations corresponding to the largest residuals (considered to be outliers) until the (24) is satisfied.

*C. Protection Level Estimation*

Even if the Chi-squared test is passed, one or more outliers may still remain in the measurements and the existence of the potential outliers is denoted by the $P_{fa}$. Therefore, the error of the localization results based on the surviving measurements mainly comes from two factors: (1) the potential missed outliers described by the probability $P_{fa}$. (2) the Gaussian noise associated with the surviving measurements.

The general expression of PL can be expressed as

$$\mathbf{PL} = \mathbf{P}_b + \mathbf{P}_n, \quad (27)$$

In particular, the first term is used to protect the bias-induced error in localization, and the second term is to bound the noise-induced error. Meanwhile, this paper aims at considering the case where multiple outliers are contained in observations even after the Chi-square test. Referring to RAIM and [40], we divide the translation into the x, y, and z and rotation into the roll, pitch, and yaw when calculating the PL in the proposed method, and then use an index vector formed by 0 and 1 to represent a state variable, for example

$$\begin{aligned} x: \mathbf{H}_1 &= [1,0,0,0,0,0]; roll: \mathbf{H}_4 = [0,1,0,0,0,0] \\ y: \mathbf{H}_2 &= [0,0,1,0,0,0]; pitch: \mathbf{H}_5 = [0,0,0,1,0,0]. \quad (28) \\ z: \mathbf{H}_3 &= [0,0,0,0,1,0]; yaw: \mathbf{H}_6 = [0,0,0,0,0,1] \end{aligned}$$

In terms of the first term in PL in (27), the bias vector for the $i^{th}$ state can be given based on error propagation and (19) [33]:

$$b_i^* = \mathbf{H}_i (\mathbf{J}^\top \mathbf{W} \mathbf{J})^{-1} \mathbf{J}^\top \mathbf{W} b, \quad (29)$$

so the norm of the bias vector is

$$\begin{aligned} \|b_i^*\| &= \sqrt{b^\top \mathbf{D} b} \\ \mathbf{D}_i &= \mathbf{W} \mathbf{J} (\mathbf{J}^\top \mathbf{W} \mathbf{J})^{-1} \mathbf{H}_i^\top \mathbf{H}_i (\mathbf{J}^\top \mathbf{W} \mathbf{J})^{-1} \mathbf{J}^\top \mathbf{W} \end{aligned} \quad (30)$$

For the multiple-outliers case, assuming there are n measurements in total, r of them contain biases, and $1 \leq r \leq (n-m)$, for example $n=7, r=2$, an index matrix $A_j$ of size $n \times r$ can be defined as

$$\mathbf{A}_j = \begin{bmatrix} 1 & 0 & 0 & 0 & 0 & 0 & 0 \\ 0 & 1 & 0 & 0 & 0 & 0 & 0 \end{bmatrix}^\top, \quad (31)$$

which means the first and second measurements have biases, the $j$ means these two measurements can be arbitrary. The matrix $A_j$ can be generated by a 2-combination of the 7 measurements denoted by $\binom{2}{7}$. Then, the biases in observations can be selected as

$$b_{ij} = \mathbf{A}_j b_i, \quad (32)$$

To make sure the PL to bound the error caused by biases, the maximum bias-induced error that results from the example $n=7, r=2$ needs to be calculated. However, since the Chi-squared test has been passed, the condition $b^T S b \leq TD$ also needs to be satisfied [33]. Based on (30) and (32), a constrained optimization model can be constructed as

$$\begin{aligned} \mathbf{P}_{b_i} &= \max b_{ij}^* = \sqrt{b_{ij}^\top \mathbf{D} b_{ij}} \\ &= \max_{\mathbf{A}_j \in \mathbf{A}_r, b_i} \sqrt{b_i^\top \mathbf{A}_j^\top \mathbf{D}_i \mathbf{A}_j b_i}, \quad (33) \\ s.t. \, b_i^\top & \mathbf{A}_j^\top \mathbf{S} \mathbf{A}_j b_i = \Gamma \end{aligned}$$

where $\Gamma = TD$, S and $D_i$ can be found in (23), (30). According to [33], this optimization question can be equally simplified by linear algebra theory as follows,

$$\max b_{ij}^* = \max_{\mathbf{A}_j \in \mathbf{A}_r} \sqrt{\Lambda_{\max}(\mathbf{A}_j, \mathbf{D}_i, \mathbf{S})\Gamma}, \quad (34)$$

where, $\Lambda_{max}(A_j, D_i, S)$ is the largest eigenvalue of

$$\mathbf{A}_j^\top \mathbf{D}_i \mathbf{A}_j (\mathbf{A}_j^\top \mathbf{S} \mathbf{A}_j)^{-1}. \quad (35)$$

By iterating through all $A_j$ and comparing the corresponding largest eigenvalue of (35), we can obtain the maximum bias-induced error in this direction.

On the other hand, the noise-induced error can be computed by

$$\mathbf{P}_{n_i} = k_i \sqrt{[(\mathbf{J}^\top \mathbf{W} \mathbf{J})^{-1}]_{i,i}}, \quad (36)$$

where $k$ is the number of the standard deviation. Empirically, the $k$ is set to 3, the so-called three-sigma rule of thumb (or $3\sigma$ rule), the corresponding probability of the values that lie within the $3\sigma$ interval is 99.73% [52]. Combining (34) and (36), the calculation of PL can be written by

$$\mathbf{PL}_i = \max_{\mathbf{A}_j \in \mathbf{A}_r} \sqrt{\Lambda_{\max}(\mathbf{A}_j, \mathbf{D}_i, \mathbf{S})\Gamma} + k_i \sqrt{[(\mathbf{J}^\top \mathbf{W} \mathbf{J})^{-1}]_{i,i}}. \quad (37)$$

In response to the line feature-based visual localization problem in this paper, we choose the two endpoints from a set of line correspondence as observations, so the index matrix $A_j$ on (31) will be set as

$$\mathbf{A}_j = \begin{bmatrix} 1 & 1 & 0 & 0 & 0 & 0 & 0 & 0 & 0 & 0 & 0 & 0 \\ 0 & 0 & 1 & 1 & 0 & 0 & 0 & 0 & 0 & 0 & 0 & 0 \end{bmatrix}^\top. \quad (38)$$

The number of observations is doubled, from 7 to 14, and the two endpoints on the same set of line correspondence are consistent.

## V. EXPERIMENT RESULT & DISCUSSIONS

The proposed framework is evaluated in real-world environments using both the public indoor dataset for the UAV and the outdoor dataset collected by UGV. We compare the performance of the proposed method with other state-of-the-art methods to evaluate the proposed optimization model and the safety-quantification method. The root means squared error (RMSE) of absolute trajectory error (ATE) [53] is the evaluation criterion of algorithm accuracy. All experiments are implemented on a desktop with Intel-Core i9-12900K and Ubuntu 20.04. The proposed framework is developed by the robot operating system (ROS) [54] and C++ language.

In terms of the evaluation of the localization performance, the algorithms compared include:
(1) ***VINS***: the VINS framework developed in [30],
(2) ***Benchmark***: the line-to-line visual localization method developed in [29].
(3) ***PPL***: the proposed visual localization method without outlier rejection.
(4) ***PPL-OR***: the proposed visual localization method with outlier rejection.

To evaluate the proposed safety-quantification model, the estimated PL is compared with the exact localization error. It is certified as safe if the estimated PL could effectively account for the potential error of the localization solution. The comparison methods include:
(1) ***3σ***: a typical method used to represent the uncertainty of the state, the computational model can be seen in (36).
(2) ***PL***: the proposed safety-quantification model.

### A. UAV Indoor Experiment

#### 1) Dataset & Comparison Methods

The EuRoc MAV (Micro Aerial Vehicle) dataset [55] is selected for evaluation, which is collected by global shutter stereo cameras (Aptina MT9V034, WVGA burri2016euroc, 20 FPS), synchronized IMU unit (ADIS16448, 200 Hz). The 3D point cloud prior maps are constructed by a Leica MS50 with an accuracy of about 1mm, and the ground-truth states are obtained by a Vicon (100 Hz) and Leica MS50. The MAV can be seen in Fig. 1. In particular, our current algorithm applies the direct output of VINS without loop-closure for state initialization, which is a completely loosely coupled form.

TABLE I: RMSE [48] of ATE (m)

| Sequence | VINS [2] | Benchmark [32] | PPL [ours] | PPL-OR [ours] |
|---|---|---|---|---|
| V1_01_esay | **0.098** | 0.164 | 0.153 | 0.152 |
| V1_02_medium | 0.110 | 0.152 | **0.092** | 0.099 |
| V1_03_difficult | 0.189 | 0.217 | 0.186 | **0.161** |
| V2_01_esay | **0.096** | 0.192 | 0.185 | 0.166 |
| V2_02_medium | 0.167 | 0.281 | 0.155 | **0.136** |
| V2_03_difficult | 0.327 | 0.441 | 0.319 | **0.312** |

RMSE of ATE: The root mean squared error of absolute trajectory error.

#### 2) Verification on Optimization Model & Outlier Rejection

The RMSE of ATE results is shown in Table I. It can be seen that the proposed methods outperform other methods in most cases, particularly in complex scenes. The improvement of ours in easy scenes with VINS is similar. However, in complex scenes (e.g., with blurred image texture), the superiority of line features can be shown. Besides, the trajectories of the *V1_02_medium* and *V1_03_difficult* are shown in Fig. 5. It can be seen that the proposed algorithm could improve the trajectory accuracy to some extent, but due to the loosely coupled structure, the state estimation results are dependent on the initial values provided by VINS, which makes the prposed algorithm has limited enhancements. On the one hand, the improved localization accuracy compared with the VINS reveals that the proposed line-based localization method can mitigate the drift from the VINS. Also, from Table I, the accuracy obtained based on the proposed optimization model is always higher than the benchmark algorithm, which indicates that the proposed line observation model could obtain better localization accuracy compared with the work in [29].

In addition, the proposed method with outlier rejection (PPL-OR) has a generally better performance than PPL, which reflects the effectiveness of the Chi-squared test-based outlier rejection algorithm used in this paper. Details about the OR efficiency are shown in Fig. 6, which displays the variation of WSSE of five keyframes as the outliers are detected and removed by a greedy algorithm sequentially. The x-axis is the number of outliers removed and the y-axis represents the residual of the observation model. By sequentially rejecting the visual outliers, the WSSE decreases.

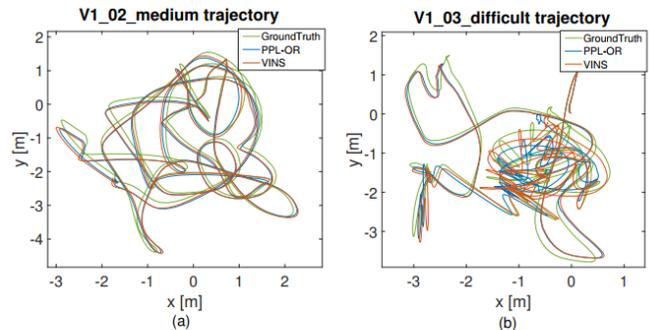

Fig. 5: (a) Trajectory comparison of V1_02_medium. (b) Trajectory comparison of V1_03_difficult.

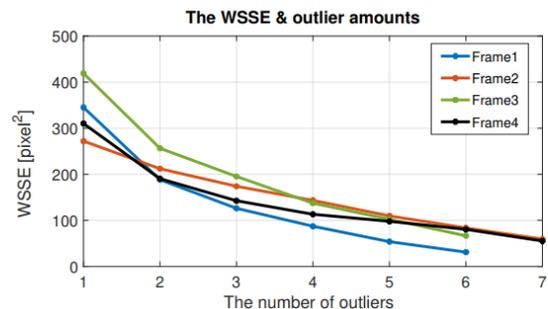

Fig. 6: The plot of the relevance WSSE and the number of outliers. The 4 frames are selected from the V1_02_medium dataset. As the outliers are detected and excluded in turn, the residual gradually decreases.

#### 3) Verification on PL

This subsection presents a protection level-based safety quantitative analysis by taking the classic V1_02_medium sequence as an example. In this experiment, we compare the bound rate of PL and $3\sigma$ (36) on state estimation errors, the bound rate of $b_r$ can be computed as

$$b_r = \frac{n}{m} \quad (39)$$

where $n$ is the number of $PL/3\sigma \geq error$, $m$ is the number of all frames. If the estimated PL could bound the actual error of the localization solution, it indicates that the proposed safety quantification is effective.

The results of translation and rotation are shown in Fig. 7 and Fig. 8, respectively, expressed by *XPL*, *YPL*, *ZPL*, *RollPL*, *PitchPL*, and *YawPL*. In Fig. 7, it is clear that PL performs much better than $3\sigma$ for error bounding, with PL having more than 70% bound rate in all three directions, while $3\sigma$ is only about 20% (see Table IV). And it can be seen that PL has a similar trend to the error in most cases. In Fig.8, PL quantifies the error significantly better than $3\sigma$, but its performance is uneven in three directions, especially in yaw rotation. The main reasons for this phenomenon are the limitation of the line features and the IMU sensor. In terms of line features, since there are many horizontal lines in space, when the object is rotated around z-axis to produce yaw rotation, the motion will be parallel to the lines making the motion useless for feature matching and state optimization, which is one of the most common degradation symptoms of line features [28].

TABLE II: The $b_r$ (%) of PL / $3\sigma$ with error using different number of outliers in translation and rotation. The covariance is 7-pixel$^2$.

| Num. | 1 | 2 | 3 |
|---|---|---|---|
| $x$ | 71.6%/ 19.6% | 74.2%/ 19.6% | 74.2%/ 19.6% |
| $y$ | 75.7%/ 24.8% | 76.8%/ 24.8% | 76.8%/ 24.8% |
| $z$ | 71.3%/ 22.9% | 73.5%/ 22.9% | 73.5%/ 22.9% |
| $roll$ | 43.9%/ 20.8% | 65.2%/ 20.8% | 65.2%/ 20.8% |
| $pitch$ | 66.6%/ 31.0% | 85.3%/ 31.0% | 85.3%/ 31.0% |
| $yaw$ | 8.23%/ 1.99% | 18.8%/ 1.99% | 18.8%/ 1.99% |

TABLE III: The $b_r$ (%) of PL / $3\sigma$ with error using different covariance assumption in translation and rotation. The number of outliers is two.

| Cov. | 3-pixel$^2$ | 5-pixel$^2$ | 7-pixel$^2$ |
|---|---|---|---|
| $x$ | 42.5%/ 7.51% | 59.8%/ 13.5% | 74.2%/ 19.6% |
| $y$ | 47.1%/ 10.0% | 64.2%/ 18.3% | 76.8%/ 24.8% |
| $z$ | 48.0%/ 11.8% | 61.3%/ 16.0% | 73.5%/ 22.9% |
| $roll$ | 36.5%/ 11.0% | 51.8%/ 15.3% | 65.2%/ 20.8% |
| $pitch$ | 60.3%/ 14.8% | 74.8%/ 22.5% | 85.3%/ 31.0% |
| $yaw$ | 6.03%/ 0.81% | 10.9%/ 1.53% | 18.8%/ 1.99% |

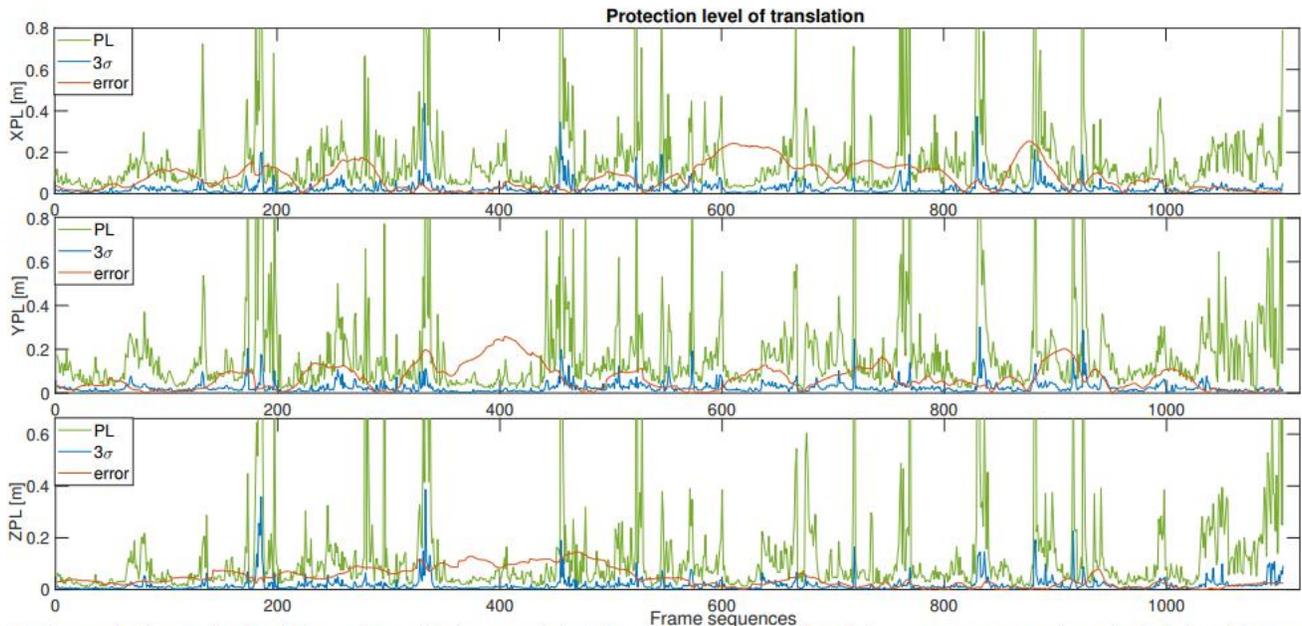

Fig. 7: The trend of protection level, $3\sigma$, and error in three translation directions. The assumption of the measurement covariance is 7-pixel and the number of outliers is set to two.

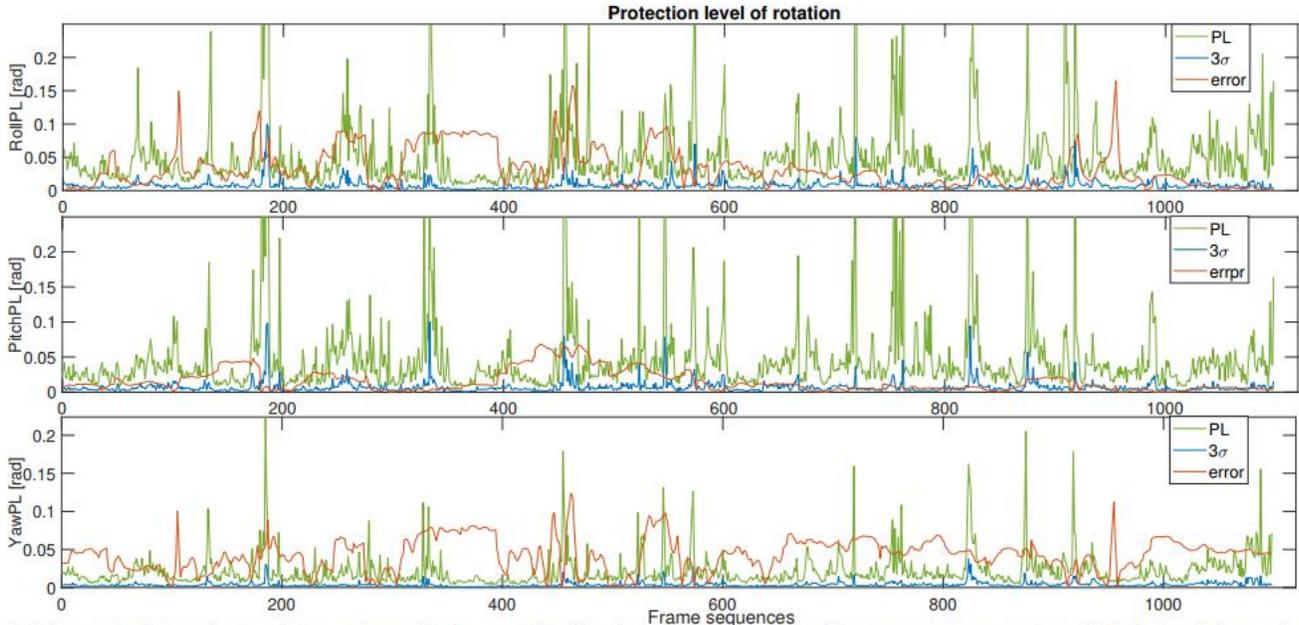

Fig. 8: The trend of protection level, $3\sigma$, and error in three rotation directions. The assumption of the measurement covariance is 7-pixel and the number of outliers is set to two.

*4) The Discussion of PL*

Ideally, the PL should be able to cover the error, but the facts contradict this. Based on the derivation of PL in Section IV-C, the calculation of PL is affected by bias-induced error resulting from outliers and noise-induced error because of observation noises. Firstly, the number of outliers existing in the current observations after passing the Chi-square test is unknown. Secondly, the noise covariance matrix of the visual measurements cannot be obtained exactly which is given empirically in this paper. All these uncertainties are factors that can affect the performance of PL estimation.

To gain more insight into PL, we evaluate the bound rate of PL and $3\sigma$ under different assumptions with multiple outliers and noise covariances in Table II. From Tables II, it can be seen that increasing the number of outliers can improve the bound rate of PL, but when the number of outliers is set to 3, it will not change, indicating that the largest number of outliers in all sequences is 2. In Table III, the measurement noise of the line feature is set to 3, 5, and 7 pixels, respectively. In practice, the measurement noise of a single pixel point feature is usually set between 1-3 pixels [56]. Since this paper uses line features, the detection of lines will introduce more errors than point features. Also, the 3D line features extracted from the point cloud map could impose observation error, so the covariance assumption is amplified here. Similarly, the same assumption is used for the Chi-squared test above. It can be seen from Table III that as the covariance increases, the rate rises, which means that PL is getting larger. In summary, more outliers and larger covariance increase PL, which is consistent with the definition of PL for quantifying safety, where larger error noise could lead to larger PL.

A specific analysis is done for the case of abnormal PL where the PL cannot reliably bound the localization error. Based on the experiments above, it is known that PL is influenced by the quality of observations. Moreover, the distribution of the line features utilized for the localization is the other sources affecting the PL estimation. It is expected that the features are uniformly distributed horizontally and vertically which could introduce optimal constraints. However, the case could occur where the features flock together which can lead to the degeneration of the state estimation [57]. To quantify the impacts of the feature distribution on the localization performance, the inverse condition number (ICN) is introduced in [57]. In particular, the ICN is determined by the inverse of the ratio between the minimum and maximum eigenvalues of $J^T J$, where, the J denotes the Jacobian matrix of the visual line constraint which is rigorously derived in the appendix of this paper.

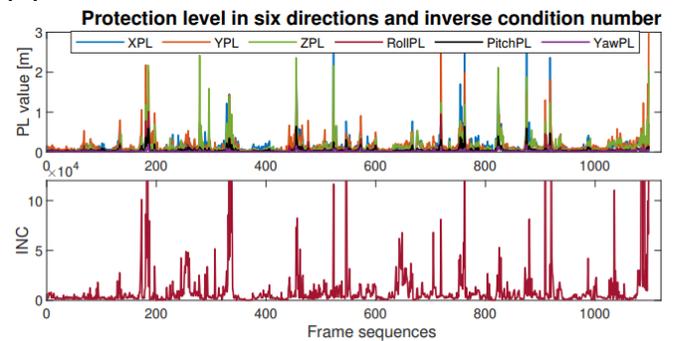

Fig. 9: The inverse condition number (ICN) and comparison of protection level, $3\sigma$, and error in three translation directions. The assumption of the measurement covariance is 7-pixel and the number of outliers is set to two.

Fig. 9 displays the ICN and PL in three directions. As seen from the figure that PL has a highly consistent trend with ICN at the abnormal mutation, which indicates that these frames have ill-conditional constraints and feature degeneration. Based on this discovery, we can conclude that feature degeneration largely causes exceptional maxima peak in PL. On the other hand, we could utilize PL to remind the feature degeneration phenomenon during state estimation to avoid serious optimization failures. Fig. 10 shows two typical feature degeneration scenarios. It is clear that most of the features in images are parallel, which makes the constraint with its vertical direction very weak, resulting in an ill-condition case.

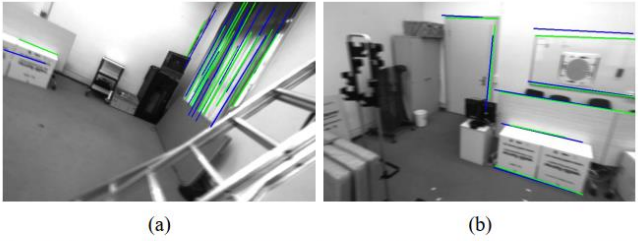

Fig. 10: Two examples of the feature degeneration. The blue and green lines express the feature matching pairs.

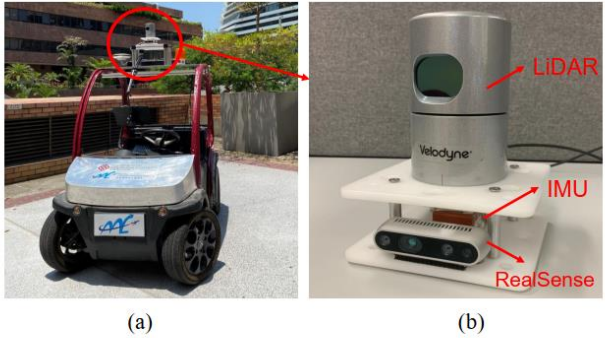

Fig. 11: (a) The UGV platform. (b) Sensor unit including LiDAR, IMU and RealSense camera.

### B. UGV Outdoor Experiment

#### 1) Experiment Setup

The UGV platform used in this experiment is shown in Fig. 11 (a), and (b) is the sensor suite, which contains a LiDAR (HDL 32E Velodyne, 10 Hz), an IMU (Xsens Mti 10, 200 Hz), and a RealSense camera (D435i, 30 Hz). The dataset is collected by driving around a square on the Hong Kong Polytechnic University (PolyU) campus (Fig.12 (b)). The total path length is 295.5 m. The IMU data and images captured by RealSense are the input of VINS. The data from LiDAR and IMU are processed by LIO_SAM implementation [58], and used to build a prior point cloud map shown in Fig. 12 (a), the origin of this 3D map is defined at the starting position of the IMU. The trajectory estimated by LIO_SAM is regarded as the ground truth of localization. With the help of the LiDAR loop closure, centimeter-level accuracy can be obtained from the LIO_SAM in the evaluated scene. Accordingly, similar evaluation criteria and comparisons with UAV experiment are adopted in this section.

#### 2) System Evaluation

Table IV lists the RMSE of ATE comparison results of the proposed method with VINS and benchmark, which can be seen that our algorithm can also improve the localization accuracy in the outdoor environment using UGV. Fig. 13 presents the estimation errors of translation and rotation, it is clear our algorithm is able to reduce the drift problem of VINS to some extent, which is evident in the z and yaw motions, while the errors in other directions can be offset periodically because the UGV trajectory is a constantly repeating circular motion. However, as the path keeps getting longer and the VINS drift increases, the effectiveness of our algorithm for removing the cumulative error gradually diminishes (see Fig. 13) due to the fact that the proposed method uses the output of VINS as the initial guess.

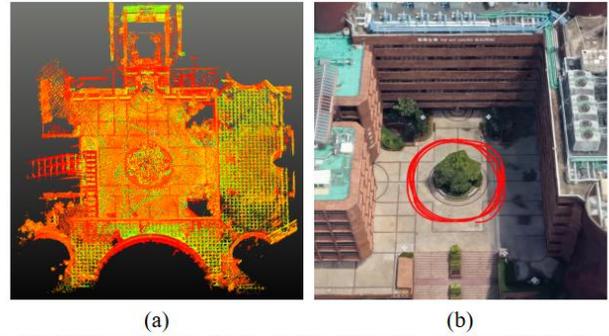

Fig. 12: (a) The regional point cloud of the Hong Kong Polytechnic University. (b) The corresponding Google 3D map of the same area. The red line indicates the trajectory of the UGV during the testing.

TABLE IV: RMSE [53] of ATE (m)

| Sequence | VINS [6] | Benchmark [30] | PPL [ours] | PPL-OR [ours] |
|---|---|---|---|---|
| PolyU Campus | 1.688 | 2.191 | 1.607 | **1.602** |

RMSE of ATE: The root mean squared error of absolute trajectory error.

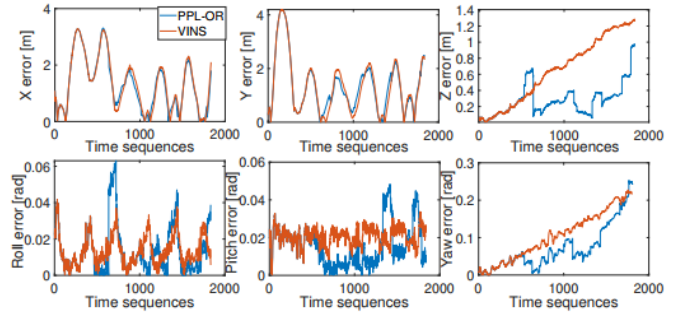

Fig. 13: The position and orientation error of PPL-OR compared with VINS.

TABLE V: The $b_r$ (%) of PL with error using different number of outliers in translation and rotation with covariance is 7-pixel

| Num. | $x$ | $y$ | $z$ | roll | pitch | yaw |
|---|---|---|---|---|---|---|
| 1 | 83.1% | 66.2% | 50.8% | 64.6% | 53.9% | 5.13% |
| 2 | 85.1% | 69.2% | 54.9% | 66.7% | 56.4% | 5.64% |
| 3 | 85.1% | 69.2% | 54.9% | 66.7% | 56.4% | 5.64% |

Based on the discussion of PL in the previous subsection, the safety quantification of UGV is verified by adjusting the number of outliers and the covariance of measurements. By fixing the covariance to 7-$pixel^2$, the relationship between the bound rate of PL and the number of outliers is shown in Table V. It is clear that the number of residual outliers in the UGV dataset is two, which is the same as the UAV experiment. Next, by fixing the number of outliers to 2 and changing the covariance from 3-$pixel^2$ to 15-$pixel^2$, the trends of the bound rate of the PL in six directions are shown in Fig. 14. As seen from the figure, the bound rates of the PL in six directions gradually level off when covariance is equal to 13.1-$pixel^2$ (the black line). In particular, the bound rate in x-direction is infinitely close to 95% (the red dashed line), which is consistent with the probability $(1-\alpha)$ set in the outlier rejection section. Based on this figure, the final covariance in the UGV experiment is set to 13.1-$pixel^2$, and the corresponding bound rates of PL and $3\sigma$ are shown in Table VI. Fig. 15 and Fig. 16 are the trends of PL, $3\sigma$, error, and the ICN in translation and rotation, respectively.

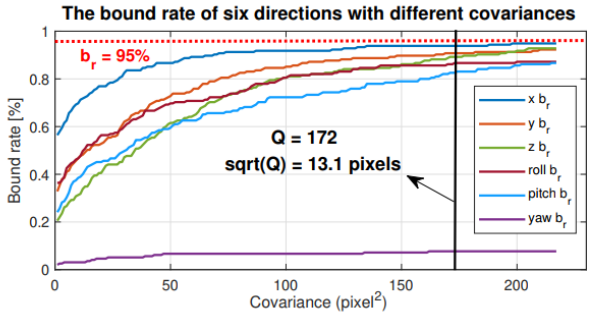

Fig. 14: The bound rate ($b_r$) of six directions with different measurement noises from 3-pixel to 15-pixel. The red dash line means the $b_r = 95\%$. The covariance corresponding to the black line is 13.1-$pixel^2$.

TABLE VI: The $b_r$ (%) of PL / $3\sigma$ under six transformation directions with covariance is 13.1-$pixel^2$ and outliers are 2.

|  | $x$ | $y$ | $z$ | $roll$ | $pitch$ | $yaw$ |
|---|---|---|---|---|---|---|
| PL | 93.9% | 89.7% | 88.2% | 85.6% | 80.5% | 7.69% |
| $3\sigma$ | 40.0% | 16.4% | 6.67% | 20.5% | 12.8% | 0.51% |

*3) Discussion*

In terms of accuracy improvement, the proposed method can eliminate the cumulative drift of VINS to a certain extent by using prior map-based localization. However, because of the loosely coupled form with VINS, the proposed method is greatly influenced by the initial guess of the state from the VINS, and when the drift is severe, our method will have difficulty converging (Fig. 13). Furthermore, the measurement noise of the line feature in the UGV dataset is about 13 pixels, which is even greater than the noise in UAV experiment. Excluding the noise of 2D line feature detection in images, the 3D line detection in point cloud map could induce more errors than UAV experiment because of the low-precision point cloud. It follows that the denseness and precision of the point cloud map is also an important factor affecting the performance of the algorithm and the PL calculation, as it impacts the line detection and the error model.

Meanwhile, same as the UAV experiment, the performance of PL is particularly poor at the yaw rotation. Comparing the localization error and PL of the yaw axis with the roll and pitch axes in Fig. 8 and Fig.16, it is obviously that the overall error of the yaw axis is larger than the others, and the PL is indeed smaller. Putting aside the small PL due to the degeneration of the line feature as known from the previous section, we will focus on the affect of IMU on the yaw localization error. Since the yaw axis of IMU is difficult to be constrained by gravity as the roll and pitch axes, the IMU will directly introduce more errors in the yaw measurement. Especially in this UGV experiment, the small radius circular motion is a severe measurement environment for the IMU, which will excite the z-axis error of gyroscope in IMU and make the yaw axis present a divergence trend . And the error of the yaw axis in Fig. 13 is consistent with this phenomenon. However, this measurement error caused by the limitation of IMU is no longer applicable to the PL calculation, as the derivation of PL is based on the defined error model of line features, which dose not take into account the IMU measurement error. In summary, the above-mentioned reasons are leading to the poor performance of PL on the yaw axis.

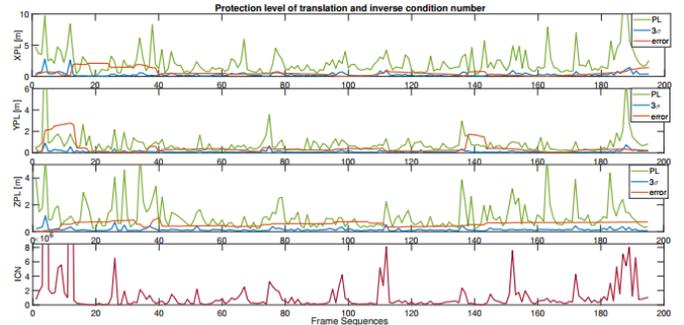

Fig. 15: The ICN and comparison of protection level, $3\sigma$, and error in three translation directions. The assumption of the measurement covariance is 13.1-$pixel^2$ and the number of outliers is set to two.

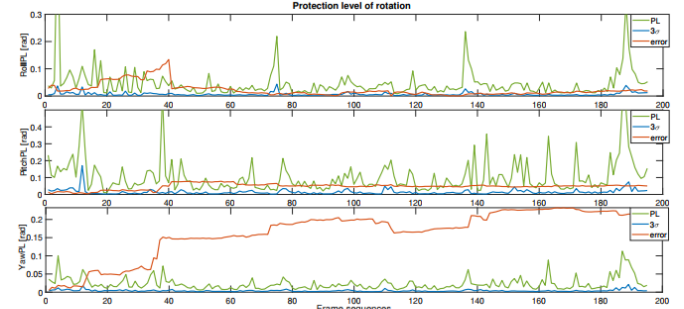

Fig. 16: The trend of protection level, $3\sigma$, and error in three rotation directions. The assumption of the measurement covariance is 13.1-$pixel^2$ and the number of outliers is set to two.

## VI. CONCLUSION AND FUTURE WORK

This paper presents a safety-quantifiable line feature-based monocular visual localization framework with point cloud maps. A new point-to-point line constrained optimization model is built to achieve accurate and robust localization in point cloud maps to eliminate cumulative drift occurring to relative positioning. A safety quantification strategy motivated by RAIM system is introduced into the current autonomous systems to evaluate the maximum value of the state estimation errors. The experimental results on UAV and UGV platforms show the proposed method outperforms the comparison algorithms in accuracy and robustness, and the error quantification of PL for state estimation is efficient in the multiple outlier assumption as well as in the translational and rotational motions.

In the future, the line feature-based optimization model will be put into VINS as a new optimization factor to build a tightly coupled schema to alleviate the dependency of the current framework on the initial solution provided by VINS. At the same time, we will also further promote the effectiveness of safety quantification from different perspectives, such as constructing a more appropriate dynamic error model by considering the effects of lighting changes, motion blur, etc., or dealing with the feature degeneracy case under the premise that PL is correlated with this situation to make the PL as close to the localization error as possible while still bounding the error so that the safety alarm mechanism can be introduced more reasonably.

## ACKNOWLEDGMENT

This research is supported by the University Grants Committee of Hong Kong under the scheme Research Impact Fund on the project R5009-21 "Reliable Multiagent Collaborative Global Navigation Satellite System Positioning for Intelligent Transportation Systems". The authors would like to thank Yihan Zhong for his help and support with the dataset collection of UGV.

## APPENDIX

### THE DERIVATION OF JACOBIAN MATRIX

The analytical expression of the Jacobian matrix of the error function e with respect to state variables is derived in this section. We have known the Jacobian can be computed by chain rule, like this

$$\frac{\partial \mathbf{e}}{\partial \delta \xi} = \lim_{\delta \xi \to 0} \frac{\mathbf{e}(\delta \xi \oplus \xi) - \mathbf{e}(\xi)}{\delta \xi} = \frac{\partial \mathbf{e}}{\partial \mathbf{p}} \frac{\partial \mathbf{p}}{\partial \mathbf{P}'} \frac{\partial \mathbf{P}'}{\partial \delta \xi}. \quad (39)$$

And we need to solve each partial derivative of this equation, the first item based on (6) and (9) can be computed by

$$\frac{\partial \mathbf{e}}{\partial \mathbf{p}} = \begin{bmatrix} \frac{\partial \mathbf{e}}{\partial \mu} & \frac{\partial \mathbf{e}}{\partial \nu} \end{bmatrix}$$
$$= \frac{2}{A^2 + B^2} \begin{bmatrix} A^2(\mu^f - \mu) + AB(\nu^f - \nu) \\ AB(\mu^f - \mu) + B^2(\nu^f - \nu) \end{bmatrix}^T. \quad (40)$$

For the second item, according to (7) and (14), $P'$ is a projected point in the camera frame and its pixel location $p(\mu,\nu)$ on the image plane is

$$s\mathbf{p} = \mathbf{KP}', \quad (41)$$

where,

$$\mu = f_x \frac{X'}{Z'} + c_x, \quad \nu = f_y \frac{Y'}{Z'} + c_y, \quad (42)$$

$f_x, f_y, c_x$ and $c_y$ is the intrinsic parameters of the camera, and then,

$$\frac{\partial \mathbf{p}}{\partial \mathbf{P}'} = -\begin{bmatrix} \frac{\partial \mu}{\partial X'} & \frac{\partial \mu}{\partial Y'} & \frac{\partial \mu}{\partial Z'} \\ \frac{\partial \nu}{\partial X'} & \frac{\partial \nu}{\partial Y'} & \frac{\partial \nu}{\partial Z'} \end{bmatrix} = -\begin{bmatrix} \frac{f_x}{Z'} & 0 & -\frac{f_x X'}{Z'^2} \\ 0 & \frac{f_y}{Z'} & -\frac{f_y Y'}{Z'^2} \end{bmatrix}. \quad (43)$$

For the third part, $P' = TP$ and give $T$ a left perturbation $\Delta T = exp(\delta \xi^\wedge)$, whose Lie algebra is $\delta \xi = [\delta \rho, \delta \phi]^T$ [44], then,

$$\begin{aligned}\frac{\partial (\mathbf{TP})}{\partial \delta \xi} &= \lim_{\delta \xi \to 0} \frac{\exp(\delta \xi^\wedge)\exp(\xi^\wedge)\mathbf{P} - \exp(\xi^\wedge)\mathbf{P}}{\delta \xi} \\ &= \lim_{\delta \xi \to 0} \frac{(\mathbf{I} + \delta \xi^\wedge)\exp(\xi^\wedge)\mathbf{P} - \exp(\xi^\wedge)\mathbf{P}}{\delta \xi} \\ &= \lim_{\delta \xi \to 0} \frac{\delta \xi^\wedge \exp(\xi^\wedge)\mathbf{P}}{\delta \xi} \\ &= \lim_{\delta \xi \to 0} \frac{\begin{bmatrix} \delta \phi^\wedge & \delta \rho \\ \mathbf{0}^T & 0 \end{bmatrix}\begin{bmatrix} \mathbf{RP}+\mathbf{t} \\ 1 \end{bmatrix}}{\delta \xi} \\ &= \lim_{\delta \xi \to 0} \frac{\begin{bmatrix} \delta \phi^\wedge(\mathbf{RP}+\mathbf{t}) + \delta \rho \\ \mathbf{0}^T \end{bmatrix}}{[\delta \rho, \delta \phi]^T} \\ &= \begin{bmatrix} \mathbf{I} & -(\mathbf{RP}+\mathbf{t})^\wedge \\ \mathbf{0}^T & \mathbf{0}^T \end{bmatrix},\end{aligned} \quad (44)$$

taking out the first 3 dimensions, we have

$$\frac{\partial \mathbf{P}'}{\partial \delta \xi} = \begin{bmatrix} \mathbf{I} & -\mathbf{P}'^\wedge \end{bmatrix}, \quad (45)$$

where,

$$-\mathbf{P}'^\wedge = \begin{bmatrix} 0 & Z' & -Y' \\ -Z' & 0 & X' \\ Y' & -X' & 0 \end{bmatrix}. \quad (46)$$

Merging formula (40) (43) and (44), the final Jacobian matrix can be solved as

$$\mathbf{J} = \frac{\partial \mathbf{e}}{\partial \delta \xi} = -\frac{2}{A^2 + B^2}[J_1, J_2, J_3, J_4, J_5, J_6], \quad (47)$$

Among that,

$$\begin{aligned} J_1 &= m\frac{f_x}{Z'} \\ J_2 &= n\frac{f_y}{Z'} \\ J_3 &= -\frac{mf_x X' + nf_y Y'}{Z'^2} \\ J_4 &= -nf_y - \frac{mf_x X' + nf_y Y'}{Z'^2}Y' \\ J_5 &= mf_x + \frac{mf_x X' + nf_y Y'}{Z'^2}X' \\ J_6 &= -m\frac{f_x Y'}{Z'^2} + n\frac{f_y X'}{Z'^2}, \end{aligned} \quad (48)$$

with,

$$\begin{aligned} m &= A^2(\mu^f - \mu) + AB(\nu^f - \nu) \\ n &= AB(\mu^f - \mu) + B^2(\nu^f - \nu) \end{aligned} \quad (49)$$

## REFERENCES

[1] A. Broggi et al., "Extensive tests of autonomous driving technologies," *IEEE Transactions on Intelligent Transportation Systems,* vol. 14, no. 3, pp. 1403-1415, 2013.

[2] W. Wen et al., "Urbanloco: A full sensor suite dataset for mapping and localization in urban scenes," in *2020 IEEE International Conference on Robotics and Automation (ICRA),* 2020: IEEE, pp. 2310-2316.

[3] X. Zhou et al., "Swarm of micro flying robots in the wild," *Science Robotics,* vol. 7, no. 66, p. eabm5954, 2022.

[4] P. K. Enge, "The global positioning system: Signals, measurements, and performance," *International Journal of Wireless Information Networks,* vol. 1, no. 2, pp. 83-105, 1994.

[5] J. Bressler, P. Reisdorf, M. Obst, and G. Wanielik, "GNSS positioning in non-line-of-sight context—A survey," in *2016 IEEE 19th international conference on intelligent transportation systems (ITSC)*, 2016: IEEE, pp. 1147-1154.

[6] W. Ding, S. Hou, H. Gao, G. Wan, and S. Song, "Lidar inertial odometry aided robust lidar localization system in changing city scenes," in *2020 IEEE International Conference on Robotics and Automation (ICRA)*, 2020: IEEE, pp. 4322-4328.

[7] G. Wan et al., "Robust and precise vehicle localization based on multi-sensor fusion in diverse city scenes," in *2018 IEEE international conference on robotics and automation (ICRA)*, 2018: IEEE, pp. 4670-4677.

[8] P. J. Besl and N. D. McKay, "Method for registration of 3-D shapes," in *Sensor fusion IV: control paradigms and data structures*, 1992, vol. 1611: Spie, pp. 586-606.

[9] T. Qin, Y. Zheng, T. Chen, Y. Chen, and Q. Su, "A Light-Weight Semantic Map for Visual Localization towards Autonomous Driving," in *2021 IEEE International Conference on Robotics and Automation (ICRA)*, 2021: IEEE, pp. 11248-11254.

[10] L. Liu, H. Li, and Y. Dai, "Efficient global 2d-3d matching for camera localization in a large-scale 3d map," in *Proceedings of the IEEE International Conference on Computer Vision*, 2017, pp. 2372-2381.

[11] Y. Xu, V. John, S. Mita, H. Tehrani, K. Ishimaru, and S. Nishino, "3D point cloud map based vehicle localization using stereo camera," in *2017 IEEE intelligent vehicles symposium (IV)*, 2017: IEEE, pp. 487-492.

[12] K. Yabuuchi, D. R. Wong, T. Ishita, Y. Kitsukawa, and S. Kato, "Visual Localization for Autonomous Driving using Pre-built Point Cloud Maps," in *2021 IEEE Intelligent Vehicles Symposium (IV)*, 2021: IEEE, pp. 913-919.

[13] M. Brown, D. Windridge, and J.-Y. Guillemaut, "A family of globally optimal branch-and-bound algorithms for 2D–3D correspondence-free registration," *Pattern Recognition,* vol. 93, pp. 36-54, 2019.

[14] R. W. Wolcott and R. M. Eustice, "Visual localization within lidar maps for automated urban driving," in *2014 IEEE/RSJ International Conference on Intelligent Robots and Systems*, 2014: IEEE, pp. 176-183.

[15] A. D. Stewart and P. Newman, "Laps-localisation using appearance of prior structure: 6-dof monocular camera localisation using prior pointclouds," in *2012 IEEE International Conference on Robotics and Automation*, 2012: IEEE, pp. 2625-2632.

[16] B. Triggs, P. F. McLauchlan, R. I. Hartley, and A. W. Fitzgibbon, "Bundle adjustment—a modern synthesis," in *International workshop on vision algorithms*, 1999: Springer, pp. 298-372.

[17] T. Caselitz, B. Steder, M. Ruhnke, and W. Burgard, "Monocular camera localization in 3d lidar maps," in *2016 IEEE/RSJ International Conference on Intelligent Robots and Systems (IROS)*, 2016: IEEE, pp. 1926-1931.

[18] P. Biber and W. Straßer, "The normal distributions transform: A new approach to laser scan matching," in *Proceedings 2003 IEEE/RSJ International Conference on Intelligent Robots and Systems (IROS 2003)(Cat. No. 03CH37453)*, 2003, vol. 3: IEEE, pp. 2743-2748.

[19] X. Zuo, P. Geneva, Y. Yang, W. Ye, Y. Liu, and G. Huang, "Visual-inertial localization with prior lidar map constraints," *IEEE Robotics and Automation Letters,* vol. 4, no. 4, pp. 3394-3401, 2019.

[20] Y. Kim, J. Jeong, and A. Kim, "Stereo camera localization in 3d lidar maps," in *2018 IEEE/RSJ International Conference on Intelligent Robots and Systems (IROS)*, 2018: IEEE, pp. 1-9.

[21] M. Schreiber, C. Knöppel, and U. Franke, "Laneloc: Lane marking based localization using highly accurate maps," in *2013 IEEE Intelligent Vehicles Symposium (IV)*, 2013: IEEE, pp. 449-454.

[22] D. Cattaneo, M. Vaghi, A. L. Ballardini, S. Fontana, D. G. Sorrenti, and W. Burgard, "Cmrnet: Camera to lidar-map registration," in *2019 IEEE intelligent transportation systems conference (ITSC)*, 2019: IEEE, pp. 1283-1289.

[23] Y. Jeon and S.-W. Seo, "EFGHNet: A Versatile Image-to-Point Cloud Registration Network for Extreme Outdoor Environment," *IEEE Robotics and Automation Letters,* 2022.

[24] A. Pumarola, A. Vakhitov, A. Agudo, A. Sanfeliu, and F. Moreno-Noguer, "PL-SLAM: Real-time monocular visual SLAM with points and lines," in *2017 IEEE international conference on robotics and automation (ICRA)*, 2017: IEEE, pp. 4503-4508.

[25] J. Lu, Z. Fang, Y. Gao, and J. Chen, "Line-based visual odometry using local gradient fitting," *Journal of Visual Communication and Image Representation,* vol. 77, p. 103071, 2021.

[26] Y. He, J. Zhao, Y. Guo, W. He, and K. Yuan, "Pl-vio: Tightly-coupled monocular visual–inertial odometry using point and line features," *Sensors,* vol. 18, no. 4, p. 1159, 2018.

[27] D. G. Kottas and S. I. Roumeliotis, "Efficient and consistent vision-aided inertial navigation using line observations," in *2013 IEEE International Conference on Robotics and Automation*, 2013: IEEE, pp. 1540-1547.

[28] H. Lim, J. Jeon, and H. Myung, "UV-SLAM: Unconstrained line-based SLAM using vanishing

points for structural mapping," *IEEE Robotics and Automation Letters,* vol. 7, no. 2, pp. 1518-1525, 2022.

[29] H. Yu, W. Zhen, W. Yang, J. Zhang, and S. Scherer, "Monocular camera localization in prior lidar maps with 2d-3d line correspondences," in *2020 IEEE/RSJ International Conference on Intelligent Robots and Systems (IROS)*, 2020: IEEE, pp. 4588-4594.

[30] T. Qin, P. Li, and S. Shen, "Vins-mono: A robust and versatile monocular visual-inertial state estimator," *IEEE Transactions on Robotics,* vol. 34, no. 4, pp. 1004-1020, 2018.

[31] P. Antonante, D. I. Spivak, and L. Carlone, "Monitoring and diagnosability of perception systems," in *2021 IEEE/RSJ International Conference on Intelligent Robots and Systems (IROS)*, 2021: IEEE, pp. 168-175.

[32] T. Walter and P. Enge, "Weighted RAIM for precision approach," in *PROCEEDINGS OF ION GPS*, 1995, vol. 8, no. 1: Institute of Navigation, pp. 1995-2004.

[33] J. E. Angus, "RAIM with multiple faults," *Navigation,* vol. 53, no. 4, pp. 249-257, 2006.

[34] M. A. Fischler and R. C. Bolles, "Random sample consensus: a paradigm for model fitting with applications to image analysis and automated cartography," *Communications of the ACM,* vol. 24, no. 6, pp. 381-395, 1981.

[35] P. H. Torr and A. Zisserman, "MLESAC: A new robust estimator with application to estimating image geometry," *Computer vision and image understanding,* vol. 78, no. 1, pp. 138-156, 2000.

[36] P. Speciale, D. Pani Paudel, M. R. Oswald, T. Kroeger, L. Van Gool, and M. Pollefeys, "Consensus maximization with linear matrix inequality constraints," in *Proceedings of the IEEE Conference on Computer Vision and Pattern Recognition*, 2017, pp. 4941-4949.

[37] T.-J. Chin, Y. Heng Kee, A. Eriksson, and F. Neumann, "Guaranteed outlier removal with mixed integer linear programs," in *Proceedings of the IEEE Conference on Computer Vision and Pattern Recognition*, 2016, pp. 5858-5866.

[38] A. Das and S. L. Waslander, "Outlier rejection for visual odometry using parity space methods," in *2014 IEEE International Conference on Robotics and Automation (ICRA)*, 2014: IEEE, pp. 3613-3618.

[39] R. Mur-Artal and J. D. Tardós, "Orb-slam2: An open-source slam system for monocular, stereo, and rgb-d cameras," *IEEE transactions on robotics,* vol. 33, no. 5, pp. 1255-1262, 2017.

[40] C. Li and S. L. Waslander, "Visual measurement integrity monitoring for uav localization," in *2019 IEEE International Symposium on Safety, Security, and Rescue Robotics (SSRR)*, 2019: IEEE, pp. 22-29.

[41] C. Zhu, M. Meurer, and C. Günther, "Integrity of Visual Navigation—Developments, Challenges, and Prospects," *NAVIGATION: Journal of the Institute of Navigation,* vol. 69, no. 2, 2022.

[42] C. Zhu, C. Steinmetz, B. Belabbas, and M. Meurer, "Feature error model for integrity of pattern-based visual positioning," in *Proceedings of the 32nd International Technical Meeting of the Satellite Division of The Institute of Navigation (ION GNSS+ 2019)*, 2019, pp. 2254-2268.

[43] X. Lu, Y. Liu, and K. Li, "Fast 3D line segment detection from unorganized point cloud," *arXiv preprint arXiv:1901.02532,* 2019.

[44] X. Gao, T. Zhang, Y. Liu, and Q. Yan, "14 lectures on visual SLAM: from theory to practice," *Publishing House of Electronics Industry, Beijing,* 2017.

[45] R. G. Von Gioi, J. Jakubowicz, J.-M. Morel, and G. Randall, "LSD: A fast line segment detector with a false detection control," *IEEE transactions on pattern analysis and machine intelligence,* vol. 32, no. 4, pp. 722-732, 2008.

[46] V. S. Varadarajan, *Lie groups, Lie algebras, and their representations*. Springer Science & Business Media, 2013.

[47] R. Hartley and A. Zisserman, *Multiple view geometry in computer vision*. Cambridge university press, 2003.

[48] P. Furgale, J. Rehder, and R. Siegwart, "Unified temporal and spatial calibration for multi-sensor systems," in *2013 IEEE/RSJ International Conference on Intelligent Robots and Systems*, 2013: IEEE, pp. 1280-1286.

[49] S. Agarwal and K. Mierle, "Ceres solver: Tutorial & reference," *Google Inc,* vol. 2, no. 72, p. 8, 2012.

[50] J. Nocedal and S. J. Wright, *Numerical optimization*. Springer, 1999.

[51] F. Gustafsson, "Statistical signal processing approaches to fault detection," *Annual Reviews in Control,* vol. 31, no. 1, pp. 41-54, 2007.

[52] F. Pukelsheim, "The three sigma rule," *The American Statistician,* vol. 48, no. 2, pp. 88-91, 1994.

[53] J. Sturm, N. Engelhard, F. Endres, W. Burgard, and D. Cremers, "A benchmark for the evaluation of RGB-D SLAM systems," in *2012 IEEE/RSJ international conference on intelligent robots and systems*, 2012: IEEE, pp. 573-580.

[54] M. Quigley *et al.*, "ROS: an open-source Robot Operating System," in *ICRA workshop on open source software*, 2009, vol. 3, no. 3.2: Kobe, Japan, p. 5.

[55] M. Burri *et al.*, "The EuRoC micro aerial vehicle datasets," *The International Journal of Robotics Research,* vol. 35, no. 10, pp. 1157-1163, 2016.

[56] K. Wu, A. M. Ahmed, G. A. Georgiou, and S. I. Roumeliotis, "A Square Root Inverse Filter for Efficient Vision-aided Inertial Navigation on Mobile Devices," in *Robotics: Science and Systems*, 2015, vol. 2: Rome, Italy.

[57] J. Zhang, M. Kaess, and S. Singh, "On degeneracy of optimization-based state estimation problems," in *2016 IEEE International Conference on Robotics and Automation (ICRA)*, 2016: IEEE, pp. 809-816.

[58] T. Shan, B. Englot, D. Meyers, W. Wang, C. Ratti, and D. Rus, "Lio-sam: Tightly-coupled lidar inertial

odometry via smoothing and mapping," in *2020 IEEE/RSJ international conference on intelligent robots and systems (IROS)*, 2020: IEEE, pp. 5135-5142.

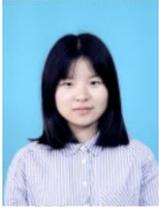

**Xi Zheng** received the B.Eng. and M.Eng degrees in astronautics from Northwestern Polytechnical University, Xi'an, China, in 2015 and 2018, respectively. She is currently working toward a Ph.D. degree with the Department of Aeronautical and Aviation Engineering, the Hong Kong Polytechnic University. Her research interests include visual SLAM and safety quantifiable localization.

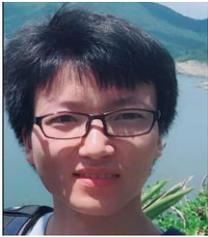

**Weisong Wen** was born in Ganzhou, Jiangxi, China. He received a Ph.D. degree in mechanical engineering, from the Hong Kong Polytechnic University. He was a visiting student researcher at the University of California, Berkeley (UCB) in 2018. He is currently a research assistant professor in the Department of Aeronautical and Aviation Engineering. His research interests include multi-sensor integrated localization for autonomous vehicles, SLAM, and GNSS positioning in urban canyons.

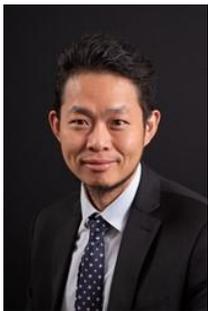

**Li-Ta Hsu** received B.S. and Ph.D. degrees in aeronautics and astronautics from National Cheng Kung University, Taiwan, in 2007 and 2013, respectively. He is currently an associate professor with the Department of Aeronautical and Aviation Engineering. The Hong Kong Polytechnic University, before he served as a post-doctoral researcher at the Institute of Industrial Science at the University of Tokyo, Japan. In 2012, he was a visiting scholar at University College London, the U.K. His research interests include GNSS positioning in challenging environments and localization for pedestrians, autonomous driving vehicle, and unmanned aerial vehicle.